\pdfoutput=1

\documentclass[11pt]{article}

\usepackage{caption}
\usepackage{subcaption}

\usepackage[final]{latex/acl}

\usepackage{times}
\usepackage{latexsym}
\usepackage{soul}

\usepackage[T1]{fontenc}

\usepackage[utf8]{inputenc}

\usepackage{microtype}

\usepackage{inconsolata}

\usepackage{graphicx}
\usepackage{amsmath}

\newcommand{\ourdataset}{ChaI-TeA}

\title{\ourdataset: A Benchmark for Evaluating Autocompletion of Interactions with LLM-based Chatbots}

\author{Shani Goren\textsuperscript{\normalfont1$\thanks{This project was done during an internship at Amazon.}$,2} Oren Kalinsky\textsuperscript{\normalfont 1}\, Tomer Stav\textsuperscript{\normalfont 1}\, Yuri Rapoport\textsuperscript{\normalfont 1}\,\, Yaron Fairstein\textsuperscript{\normalfont 1}\\\, \textbf{Ram Yazdi}\textsuperscript{\normalfont 1}\ \textbf{Nachshon Cohen}\textsuperscript{\normalfont 1}\ \textbf{Alexander Libov}\textsuperscript{\normalfont 1}\, \textbf{Guy Kushilevitz}\textsuperscript{\normalfont1}\\
\textsuperscript{1}Amazon Research \,\textsuperscript{2}Technion - Israel institute of technology \\\\ 
    {\tt\{\href{mailto:shani.goren@gmail.com}{shani.goren},
    {\href{mailto:orenkalinsky@gmail.com}{orenkalinsky},
    {\href{mailto:Yyfairstein@gmail.com}{yyfairstein}{\}@gmail.com}}}}\\
    {\tt\{\href{mailto:alibov@amazon.com}{alibov},
    {\href{mailto:stavt@amazon.com}{stavt}},
    {\href{mailto:rtu@amazon.com}{rtu}},
    {\href{mailto:ramyazdi@amazon.com}{ramyazdi}},
    {\href{mailto:nachshon@amazon.com}{nachshon}},
    {\href{mailto:guyk@amazon.com}{guyk}}\}@amazon.com}}

\begin{document}
\maketitle
\begin{abstract}
The rise of LLMs has deflected a growing portion of human-computer interactions towards LLM-based chatbots.
The remarkable abilities of these models allow users to interact using long, diverse natural language text covering a wide range of topics and styles. Phrasing these messages is a time and effort consuming task, calling for an autocomplete solution to assist users. 
We present \textbf{ChaI-TeA}: \textbf{Cha}t \textbf{I}n\textbf{te}raction \textbf{A}utocomplete; An autocomplete evaluation framework for LLM-based chatbot interactions. The framework includes a formal definition of the task, coupled with suitable datasets and metrics. We use the framework to evaluate 
$9$ models on the defined auto completion task, finding that while current off-the-shelf models perform fairly, there is still much room for improvement, mainly in ranking of the generated suggestions. We provide insights for practitioners working on this task and open new research directions for researchers in the field. We release our framework\footnote{\url{https://github.com/amazon-science/ChaiTea-chat-interaction-autocomplete}}, to serve as a foundation for future research.
\end{abstract}

\section{Introduction}

Large Language Models (LLMs) have revolutionized many NLP applications \cite{openai}. 
A prominent example is automatic chatbots; what used to be confined, topic-specific applications often requiring the user to use restricted language or choose from a closed list of interaction options, have been transformed. 
These applications, powered by LLMs, are now one-stop-shops successfully communicating in unbounded natural language while acting as experts on a wide variety of topics \cite{gpt4,gemini}. 
Due to their remarkable abilities, LLM-based chatbots differ significantly from prior human-computer communication methods. Interactions with these chatbots are usually long, unique and cover a large range of topics and language styles, using unstructured natural language. Due to this nature, users invest much time and thought in communicating their needs to the chatbot, calling for solutions to reduce their effort \cite{acforgenai}. 

\begin{figure}
    \centering
    \includegraphics[width=1\columnwidth]{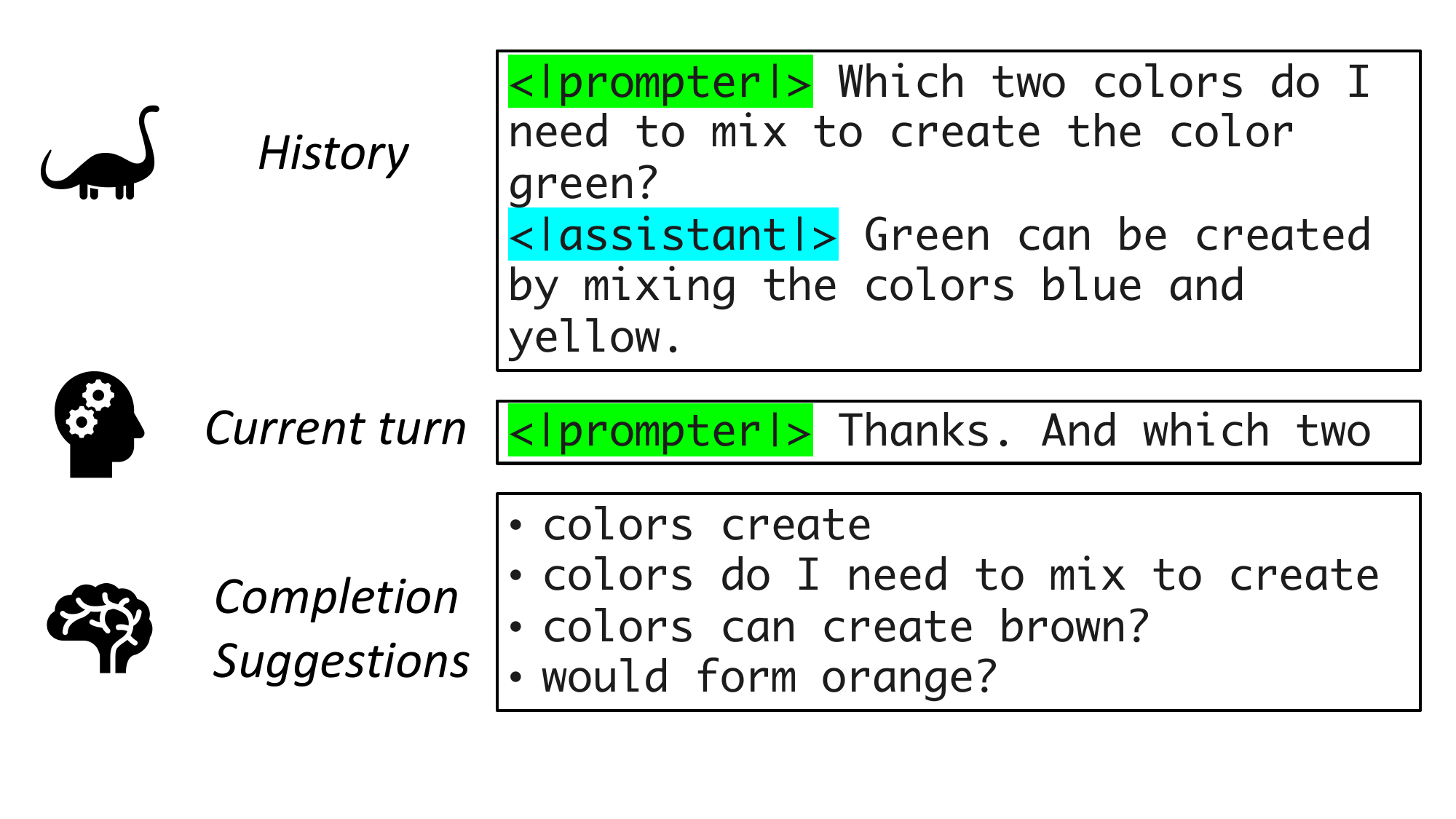}
    \caption{The chatbot interaction autocompletion task. Given the conversation history and the current turn's prefix, task is to suggest suitable completions.}
\label{fig:task_figure}
\end{figure}




AutoComplete (AC) methods have been shown to be effective in saving users' time and reducing their cognitive load in many different use-cases, suggesting that such a solution might be of value for the LLM-chatbot interaction use-case as well. 
The popular query autocomplete scenario \cite{queryACsurvey} focuses on search queries. Classic solutions often rely on recurrence, making them irrelevant for the long unique natural language text found in chatbot interactions \cite{acforgenai}. Later solutions include generative models \cite{sordoni2015hierarchical, park2017neural}, but still focus on short semi-structured queries. Code autocomplete \cite{codesurvery} deals with structured language, and often relies on the ability to run the code and check its output in order to evaluate solutions. 
Lastly, email (human-human) interactions \cite{gmail}, which bear a closer resemblance to human-chatbot interactions due to their natural language communication, also differ in several key aspects. These include the number of participants and their roles, the more formal writing style of emails and the nature of the topics discussed. In broader terms, human-human textual interactions (e.g., emails, but also texts from other kinds of messaging platforms) differ from human-chatbot interactions in the fact that human-chatbot interactions involve a human and a model-based assistant, making them more instructional and knowledge-seeking. For example, the prompts \textit{“Give me the latest updates of the war in Ukraine as of the 31st of January.”} and \textit{“Write a web scraping program in python capable of ...”} are taken from the OASST dataset used in this work to demonstrate typical examples for a human-chatbot interaction, which are highly unlikely to be found in a human-human messaging platform.

In this paper, we introduce the task of autocompleting user interactions with LLM-based chatbots. 
We present \textbf{ChaI-TeA}: \textbf{Cha}t \textbf{I}n\textbf{te}raction \textbf{A}utocomplete; A framework for evaluating autocomplete solutions for LLM-based chatbot interactions. It includes a formal definition of the task, suitable datasets tailored for autocomplete, suitable metrics, and baseline results.
We go on to highlight some valuable insights. First, we explore how performance can be traded off for lower latency, a key factor in autocomplete solutions. Second, we show that models can exploit distant history to suggest completions. Third, it is beneficial to enable completions of various lengths (as opposed to only single words or full turns). 
We highlight a key factor in improving these solutions: we find that models tend to generate completion suggestions well, but are not as good at ranking these generated suggestions. Given that users can ingest a small amount of suggestions at each turn, ranking is an important component in an offered solution. Therefore, we advocate for future research in the field to focus on this aspect.

\section{Task Definition} \label{sec:task}
The chatbot interaction completion task focuses on completing user turns in user-chatbot interactions.
Similarly to \cite{chitnis2024sequential}, we model it as a sequential task; completions are suggested at each typing step (i.e., after a user types a character).
Formally, at each step $t$, an autocomplete solution (denoted by $AC$) is given a context $C$ containing all previous conversation turns, originating from both the user and the chatbot, and the prefix of the current user turn denoted as $p_t$. 
The autocomplete solution should then return a set of $k$ completions, $c_{t_1},...,c_{t_k}$, possibly of varying lengths. 

Each completion step can be described as: \\ \centerline{$AC(C, p_t) = \{c_{t_1}, c_{t_2}, ..., c_{t_k}\}$}

After receiving the set of completions, the user can either accept a completion or continue typing. If a completion $c_{t_i}$ is accepted, the prefix is updated such that $p_{t+1} = p_t + c_{t_i}$. Then, whether the user selected a completion or continued typing, a new completion step is initiated, until reaching the end of the user's turn. A single completion step is illustrated in Figure \ref{fig:task_figure}, and full turns completions can be found in the Appendix in Table~\ref{table:full_interaction}.


\section{Experimentation}
\subsection{Datasets} \label{datasets}
\textbf{Open Assistant (OASST)} \cite{kopf2024openassistant} is a human-annotated assistant conversation corpus.
\textbf{ShareGPT}\footnote{\url{https://sharegpt.com/}, dataset version that was used: \textit{anon8231489123\/ShareGPT\_Vicuna\_unfiltered}} contains user-LLM-chatbots conversations collected by the ShareGPT API.

To curate the data for our task, we take all English conversations and for each user-turn extract all possible prefixes and pair each with the entire conversation history up to that point
as its context. The suffix of the original prompt is the ground truth completion.
Table \ref{tab:datasets} summarizes the statistics of the datasets used in our experiments.

\begin{table}[t]
  \centering
\resizebox{1\columnwidth}{!}{
  \begin{tabular}{lcccccc}
    \hline
               & \multicolumn{2}{c}{\textbf{OpenAssistant}} & \multicolumn{2}{c}{\textbf{ShareGPT}} \\
                              & \textbf{Train} & \textbf{Test}            & \textbf{Train} & \textbf{Test*} \\
    \hline
    Conversations             & 5,144          & 277                    & 88,259         & 1,190 \\
    Messages                  & 22,749         & 1,182                  & 317,536      & 1,494         \\
    Prefixes             & 536,215        & 26,394                 & 16,801,251    & 22,323        \\
    \hline
  \end{tabular}
  }
  \caption{\label{tab:datasets}
    \textbf{Dataset Statistics.} *Since ShareGPT does not include a test split, we randomly sampled one of comparable size to the OASST test set.
  }
\end{table}


\subsection{Metrics}
As solutions are allowed to propose $k$ completions at each step, metrics evaluate the performance taking $k$ into account, denoted as @$k$.

As we are looking to form a benchmark, we turn to metrics that can be computed offline. We remark that ideally, we would also like to measure the user's saved time or reduced cognitive load but doing so would require running some experiment or user study for each new proposed solution.

For simplicity, we simulate acceptances (i.e., is one of the proposed completions accepted by the user?) using exact match comparison to the ground truth user turn.

\textbf{Saved typing.} Inspired by code completion metrics \cite{codium},
our goal is to save the user typing effort. Therefore, we seek a metric that quantifies the portion of the text completed by the $AC$ solution.
While simply dividing the length of the accepted text by the length of the full turn would achieve this, this metric would not consider the number of acceptances needed to generate the accepted text. To demonstrate this issue, consider two different solutions successfully completing the full turn; the first solution does this by completing single words one by one, while the other completes the entire turn in its first attempt. The naïve metric would score the two solutions the same, although it's clear we should prefer the second solution. To mitigate this issue, we propose the following metric:
\begin{equation*}
    \text{saved@}k = 
   \frac{len(\text{accepted\_text}) - \text{\#acceptances}}{len(\text{full\_turn}) - 1}
\end{equation*}
where $len(x)$ is the number of characters in string $x$. No acceptances during the user's turn lead to a score of $0\%$ while a single acceptance completing the full turn leads to a score of $100\%$.

\textbf{Latency.} Latency is a critical factor that cannot be overlooked when assessing $AC$ solutions. Even if the completions are perfect, they are rendered useless if the user proceeds to type before receiving the suggestions. We report the mean and the $90 \textsuperscript{th}$ percentile (p90) of the inference time.





\subsection{Autocomplete Solutions} \label{models}
As our task resembles the language modeling task, a called-for solution is utilizing LMs. 
This allows us to experiment with a wide variety of models ranging in size, latency and quality, while avoiding extremely large LLMs as their latency is not feasible for this task\footnote{Generating completion suggestions with a 70B LLM takes on average 6 seconds.}.
Our evaluation encompassed a diverse set of popular LMs: Mistral-7B \cite{mistral}, Gemma-7B \cite{gemma}, Phi-3-mini \cite{phi-3}, GPT-2-XL \cite{gpt2}, Mamba \cite{mamba}, and SmolLm\footnote{\url{https://huggingface.co/blog/smollm}}. We also evaluate instruct-tuned variants of these models whenever one is available (Zephyr, Gemma)\footnote{The lack of published instruct-tuning datasets for some models prevents us from confirming the absence of data leakage. Still, our observations did not reveal any abnormal results.}.
Inference was performed on a single NVIDIA A10G GPU, taking 150 hours in total.

To generate $k$ completions from the LMs, we adopt the following procedure: we provide the model with the full context concatenated with the prompt prefix. We then use the model to generate $n_c$ completions sampled with temperature $1.0$, 
stopping when reaching EOS or after $n_t$ tokens. Since completions can vary in length, each word-prefix of a completion can also be considered as a standalone completion. Hence, this process generates up to $n_c \times n_t$ completion candidates. Finally, we choose the $k$ suggestions to present to the user by ranking the completions based on their perplexity score, computed using the LM probabilities:
\begin{equation*} 
PPL(w_1,w_2,\ldots,w_n) = e^{{-\frac{1}{n} \sum\limits_{i}^{n}{\log p(w_i|w_1,\ldots,w_{i-1})}}}
\end{equation*}



\subsection{Initiating Suggestion Generation}
Suggesting completions after each character has some downsides compared to suggesting only at an end of a word. First, as the average length of an English word is more than $4$ characters, the computational cost more than quadruples\footnote{While using caching techniques can help mitigate some of the required compute, we observe (e.g., in Fig. \ref{fig:hp}) that token generation requires a considerable computation time, that cannot be mitigated using caching.}.
Second, it has been shown that when typing, users tend to pause much longer between words than between same-word characters \cite{keys_to_writing}. This allows more room to suggest completions between words.
Third, LLMs are known to under-perform on character level tasks, since most tokenizers only use character level tokens as a fallback\footnote{For example, [DOG] is a token in most tokenizers, but given the prefix "I love my pet d", the model will likely use the character level token for [D], and the tokens [OG] or [O][G] are unlikely to be generated, since the model probably didn't encounter this token sequence during training.} \cite{llms_char}.


To compare how frequently character level suggestions are accepted compared to word level suggestions, we also tracked \emph{\textbf{acceptance rate}}: the percentage of completion steps that ended in an acceptance.



Results on the OpenAssistant validation set ($n_c=5, n_t=20$) show that mid-word suggestions degrade the acceptance rate by $\sim 60\%$ while only slightly improving saved@$k$ by $\sim 3.2\%$. Interestingly, Mamba, which uses a character-level tokenizer, behaves similarly to the other models. Full details of this experiment are reported in Appendix \ref{sec:char_vs_word}. We conclude that mid-word suggestions are rarely accepted, and do not justify their drawbacks. Additional efforts are needed to make mid-word suggestions effective, which we leave for future work. For the remainder of this paper, completion suggestions are provided only at the end of a word.
Consequently, throughout the rest of our experiments we observed that acceptance rate@$k$ is highly correlated with saved@$k$. Therefore, we exclude acceptance rate results from the main paper and present it in the appendices.

\begin{figure}
\centering
    \includegraphics[width=6.86cm]{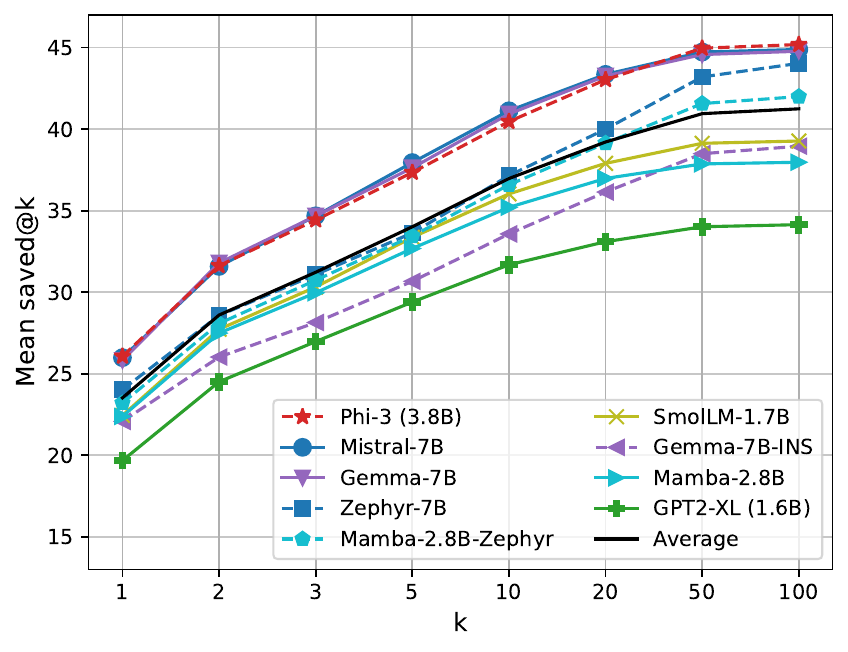}
    {\caption{\textbf{saved@\textit{k}} on OASST for varying $k$ values.}\label{fig:saved@k_vs_k}}
\end{figure}

\subsection{Benchmarking \ourdataset}
\label{subsec:benchmark}
We benchmark all models described in Section \ref{models} on both curated datasets (Section \ref{datasets}). Results on OASST for varying $k$ values are shown in Figure \ref{fig:saved@k_vs_k}. We consider $k$ values up to $100$, which encompasses all generated completions (at most, $n_c \times n_t$), to show the potential given a perfect ranking solution.
While current models are able to perform fairly on this task -- saving the user the typing of up to $45\%$ of the characters  --  there is still much room for improvement.
There is a noticeably large performance gap between small, realistic, $k$ values and larger values, suggesting that while in many cases models are able to generate the correct completion, their ranking of completions is far from perfect. In line with prior work \cite{selfcheckgpt, 
perplexity_ood, fact_checking_ppl}, we conclude that perplexity is insufficient for confidence ranking. Full benchmark results on OASST and ShareGPT can be found in Appendix \ref{app:benchmark}. 

Finally, we observe that further improvement can be gained by fine-tuning models on the AC task. Detailed results are presented in Appendix \ref{app:finetune}.

\section{Further Analysis}
\label{sec:further_analysis}

\textbf{Latency-Performance Trade-Off.}
Given the practical importance of latency in AC solutions, we explore how performance can be traded off for reduced latency. To illustrate this trade-off, we varied the previously mentioned hyperparameters $n_c$ and $n_t$, as well as the context length given to the model. We capped the conversation history concatenated with the turn prefix at different lengths, to determine whether giving the model access to the entire conversation context is both helpful and worthy of the extra latency costs.

Suggestions are offered between words, meaning that once the user begins typing the next word they become irrelevant. Hence, we find it appropriate to use the mean time between typed words -- 718 \textit{ms}, reported by \cite{keys_to_writing} -- as a benchmark. 

Results per latency budget, presented in Table \ref{table:latency}, show that it is preferable to generate more completions, while reducing the number of generated tokens and context length. Also, additional context is beneficial, suggesting that information useful for autocomplete can sometimes be found far before the end of the prefix. Results on all configurations are reported in the appendix in Table \ref{table:latency_full}.

\begin{table}[ht]
  \centering
\resizebox{1\columnwidth}{!}{
\begin{tabular}{lccccc} \hline \hline
Latency & \multicolumn{3}{c}{Best Configuration} & saved@$100$ & Latency \\ 
Budget (ms) & $n_c$ & $n_t$ & Hist. Len & & p90 (ms) \\ \hline
$<150$ & $5$ & $3$ & $50$ & $23.45$ & $148$ \\
$<300$ & $5$ & $5$ & $250$ & $38.32$ & $275$ \\
$<450$ & $5$ & $3$ & $1000$ & $41.10$ & $388$ \\
$<600$ & $5$ & $5$ & $1000$ & $44.08$ & $451$ \\
$<750$ & $5$ & $5$ & $1000$ & $44.08$ & $451$ \\
$>750$ & $5$ & $10$ & $Full$ & $45.75$ & $974$ \\
\hline \hline
\end{tabular}}
  \caption{
    \textbf{Latency-Performance Trade-Off.} Mistral-7B evaluated on 
    the OASST test set. $n_c \in \{3,4,5\}$, $n_t \in \{3,5,10,20\}$, and context length $len(C) \in \{50, 250, 1000, Full\}$ (measured in characters). In total, 48 hyper-parameter configurations were evaluated. For each latency budget, we report the configuration with the highest saved@$100$ score that fits the budget.
  }\label{table:latency}
\end{table}


\textbf{Varying completion lengths.} 
A common practice for autocomplete practitioners wanting to simplify their methods is restricting completions to single words.
The other end of this scale, also widely used, is allowing only full completions- completing until the end of the query/function/sentence. 
To this end, we compare completions of varying lengths to single word and full sentence completions to check whether allowing any-length completions improves quality. 
Average results across all models are presented In Table \ref{table:baselines_avg} (Full results can be found in Table \ref{table:baselines_full}).
$\text{saved@}k$ metric improves for $k = 100$ when allowing suggestions of varying length, indicating this can improve the user's typing experience. The fact that this is not the case for the lower $k$ values indicates, once more, that the ranking method we use (the model's perplexity) is far from ideal.

\begin{table}[h]
\label{table:main}
  \centering
\resizebox{1.0\columnwidth}{!}{
\begin{tabular}{lccc} \hline \hline     
      & \multicolumn{1}{c}{saved@1} & \multicolumn{1}{c}{saved@3} & \multicolumn{1}{c}{saved@100} \\ \hline \hline

   Single Word  & $\mathbf{24.10}$ / $\mathbf{22.28}$ & $\mathbf{31.97}$ / $28.63$ & $33.12$ / $29.52$  \\  
    Full& $12.30$ / $10.44$ & $15.91$ / $13.29$ & $16.47$ / $13.70$ \\ 
    Partial  & $23.43$ / $22.03$ & $31.21$ / $\mathbf{28.85}$ & $\mathbf{41.27}$ / $\mathbf{36.77}$ \\ 
 
 \hline
             \hline
\end{tabular}
}
  \caption{\label{table:baselines_avg}
    Average scores of partial completions vs single word and full sentences. OpenAssistant / ShareGPT.
  }
\end{table}


 


\textbf{Characteristics of completions.} We observe that different models are able to generate diverse suggestions of different lengths. Completion suggestions offered by the different models are presented in Table \ref{table:examples}. When looking at accepted completions, we see that while most acceptances are single word completions ($60\%-70\%$), the models are able to generate longer acceptances; more than $15\%$ span over $3$ words or longer. The lengths of acceptances are presented in Figure~\ref{fig:length}.



\section{Conclusions}
In this work, we showcase the task of autocompleting user interactions with LLM-based chat-bots. 
We formally define the task and design an evaluation framework, and use it to test $9$ different models.
Results show that while LMs are able to perform fairly, there is room for a tailored solution to improve upon them, especially in the ranking of completion candidates.
We show that models can exploit distant history,
that enabling
completions of different lengths is beneficial and that reducing latency for this task should be done by reducing context length and length of completions as opposed to generating less completions. We hope our framework will encourage further work in this area, which we believe
holds great potential value for users across various LLM chat-bot applications.

\section*{Limitations}
\textbf{Exact Match.} We use exact-match to simulate acceptances. While this is standard practice in autocomplete works, it may not fully represent real-world scenarios in which a user might accept a completion even if it's not the exact wording they were thinking of. Although some works use generation metrics like BLEU or ROUGE to simulate full sentence acceptances, these metrics fail to capture semantic similarity between partial completion suggestions and ground truths, making them a problematic solution because even a very high score may not represent an accept and vice versa. Moreover, it is a non-trivial task to infer what a user will accept after semantic partial matches since the text diverged from the ground truth. We evaluated using the Claude3-Sonnet model to determine whether a suggestion should be accepted or not and discovered this to be a very challenging task. Thus, we leave it for future work.


\textbf{Datasets.} Both datasets used have one significant limitation: they where collected without the presence of an autocomplete solution. It is possible that users alter their behavior when completion suggestions are presented to them. If this is true, it will not be reflected in our framework. We note that taking this into account is far from trivial, because even if data is collected in the presence of some autocomplete solution, this data will be biased towards the specific solution used in the collection process, giving an unfair advantage when judging solutions similar to it.

\textbf{Word-level completions.} Most of the results presented in this paper assume completions are only suggested at the end of words. While this is possible to achieve in a real-world scenario, it would require some component assessing whether an end of a word is reached or not. This solution will have to run online, and in short latency. Since our experiments are run offline, the full turn was available for us and we could simply check when the end of a word was reached.
\clearpage




\bibliography{latex/acl_latex}
\appendix
\newpage
\section{Character vs Word level completions} \label{sec:char_vs_word}
\begin{figure}[t]
    \includegraphics[width=7cm]{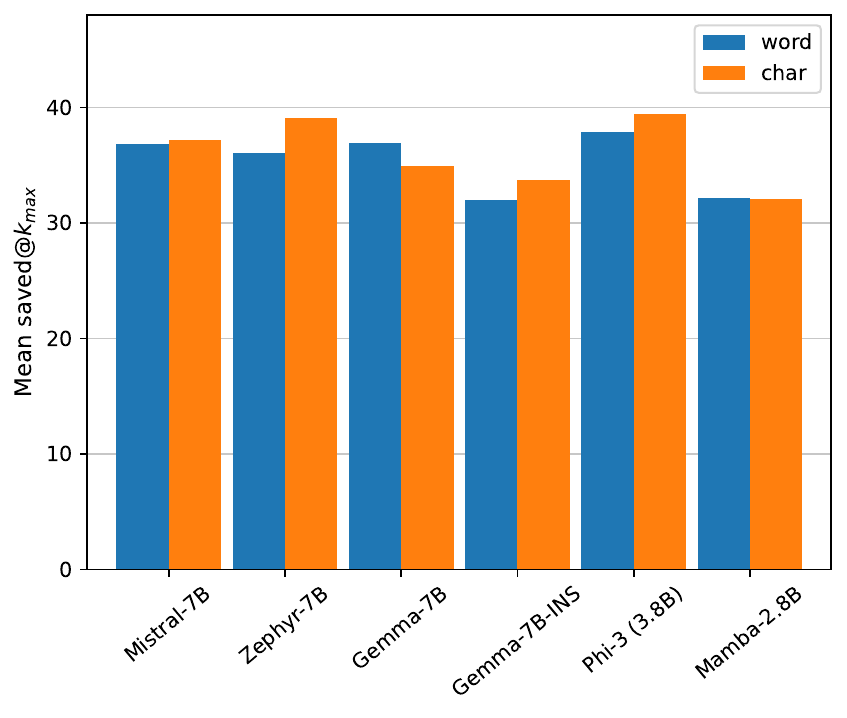}
    {\caption{saved@$k$ comparison between solutions suggesting completions after words and characters.}\label{fig:saved_word_char}}
\end{figure}

\begin{figure}[t]
    \includegraphics[width=7cm]{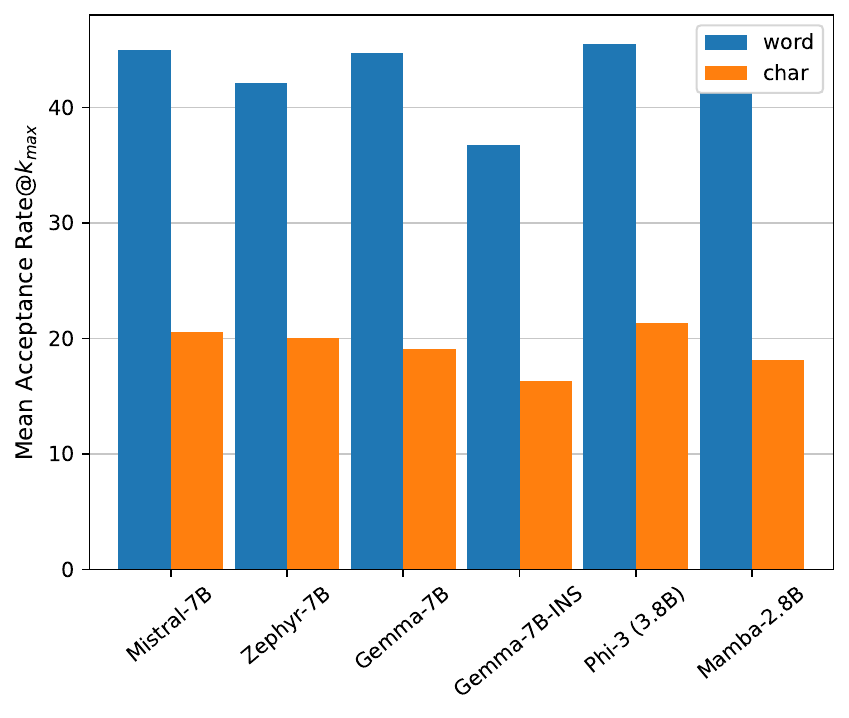}
    {\caption{\textit{acceptance rate} comparison between solutions suggesting completions after words and characters.}\label{fig:acc_word_char}}
\end{figure}

In this section we detail our comparison between suggesting completions after each character compared to doing so only at the end of words. We consider $k_{\text{max}}$, i.e., all generated completions (at most, $n_c \times n_t$). While this scenario is not realistic, since for the \emph{best} configuration it means presenting the user with $100$ completion options, it shows the potential each solution has with a perfect ranking solution.
We start with saved@$k$. Issuing suggestions after each character is expected to improve this metric compared to issuing suggestions after each word. This is due to the fact that this metric does not penalize on unaccepted suggestions. Therefore if every mid-word suggestion is ignored by the user, the metric will remain unchanged. If some mid-word suggestion are accepted, the metric is expected to rise. In Figure \ref{fig:saved_word_char} we show results on saved@$k_{\text{max}}$. Indeed, the metric is improved when suggesting after each character, but the difference is minor (on average across models, $3.2\%)$. We note that even for Mamba, which uses a character-based tokenizer, the difference is very small.
Next, we compare the same solutions on acceptance rate. Results in Figure \ref{fig:acc_word_char} show that acceptance rate for the solutions suggesting only at end of words is much higher (on average, $\sim 130\%$ improvement), suggesting that the mid-word suggestions are rarely accepted. 

\section{Full Benchmark Results} 
\label{app:benchmark}
The full results are reported in Table \ref{table:results}. For each model mentioned in Section \ref{models}, we report results for two hyper parameter combinations: \textit{best} 
is a version aimed at optimizing the quality ($n_c = 5, n_t=20$), while \textit{fast} is a version aimed at optimizing the latency ($n_c = 1, n_t=5$). A full hyper parameter study can be found in Appendix \ref{sec:hp_study}. We report on both datasets presented in Section \ref{datasets} for different $k$ values. Further analysis of the effect of $k$ can also be found in Appendix \ref{sec:hp_study}.

\begin{table*}[h]
  \centering
\resizebox{1.0\textwidth}{!}{
\begin{tabular}{llcccccccc} \hline \hline &      & \multicolumn{2}{c}{$k=1$}                                     & \multicolumn{2}{c}{$k=3$}                                     & \multicolumn{2}{c}{$k_{max}$}                                 & \multicolumn{2}{c}{Latency (ms)} \\ 
&      & \multicolumn{1}{c}{saved@1} & \multicolumn{1}{c}{acc. rate@1} & \multicolumn{1}{c}{saved@3} & \multicolumn{1}{c}{acc. rate@3} & \multicolumn{1}{c}{saved@k} & \multicolumn{1}{c}{acc. rate@k} & \multicolumn{1}{c}{mean} & \multicolumn{1}{c}{p90} \\ \hline \hline

Mistral-7B & best  & $\underline{25.97}$ / $\underline{24.67}$ & $\underline{32.23}$ / $\underline{32.32}$ & $\mathbf{34.66}$ / $\underline{32.76}$ & $\underline{37.65}$ / $\underline{38.35}$ & $\underline{44.86}$ / $\mathbf{41.04}$ & $\underline{50.56}$ / $\underline{49.33}$ & $834$ / $1479$ & $1288$ / $2485$ \\  
            & fast  & $26.23$ / $24.32$ & $32.46$ / $31.94$ & $33.29$ / $30.75$ & $36.13$ / $36.19$ & $35.02$ / $32.12$ & $38.02$ / $37.84$ & $201$ / $356$ & $313$ / $588$ \\ \hline 

Zephyr-7B & best  & $24.01$ / $23.47$ & $29.81$ / $30.22$ & $31.06$ / $29.89$ & $32.91$ / $34.31$ & $44.00$ / $40.85$ & $47.91$ / $47.65$ & $870$ / $1520$ & $1313$ / $2512$ \\  
            & fast  & $24.63$ / $23.39$ & $30.81$ / $30.48$ & $31.28$ / $29.16$ & $34.03$ / $34.30$ & $33.49$ / $31.02$ & $36.62$ / $36.63$ & $214$ / $368$ & $320$ / $589$ \\ \hline 



Gemma-7B & best  & $25.80$ / $\mathbf{24.84}$ & $\mathbf{32.34}$ / $\mathbf{32.71}$ & $\mathbf{34.66}$ / $\mathbf{32.91}$ & $\mathbf{37.72}$ / $\mathbf{38.93}$ & $44.75$ / $\underline{41.02}$ & $50.02$ / $\mathbf{49.44}$ & $961$ / $1587$ & $1423$ / $3032$ \\  
            & fast  & $25.62$ / $23.77$ & $32.48$ / $31.57$ & $32.55$ / $29.89$ & $35.64$ / $35.61$ & $34.25$ / $31.44$ & $37.93$ / $37.67$ & $239$ / $412$ & $358$ / $684$ \\ \hline 

Gemma-7B-INS & best  & $22.08$ / $21.65$ & $28.33$ / $28.72$ & $28.13$ / $27.32$ & $30.83$ / $31.78$ & $38.94$ / $35.67$ & $41.81$ / $41.24$ & $837$ / $1522$ & $1355$ / $2981$ \\  
            & fast  & $22.41$ / $21.29$ & $29.14$ / $28.70$ & $28.29$ / $26.50$ & $31.61$ / $31.72$ & $30.39$ / $28.13$ & $34.27$ / $33.99$ & $245$ / $421$ & $358$ / $702$ \\ \hline 

Phi-3 (3.8B) & best  & $\mathbf{26.07}$ / $24.18$ & $32.07$ / $31.25$ & $\underline{34.42}$ / $31.83$ & $36.76$ / $36.99$ & $\mathbf{45.18}$ / $39.91$ & $\mathbf{50.81}$ / $47.84$ & $510$ / $879$ & $786$ / $1466$ \\  
            & fast  & $26.13$ / $23.25$ & $32.26$ / $30.39$ & $33.21$ / $29.60$ & $35.91$ / $34.95$ & $34.82$ / $30.92$ & $37.98$ / $36.70$ & $117$ / $208$ & $185$ / $344$ \\ \hline 

Mamba-2.8B & best  & $22.36$ / $21.66$ & $29.44$ / $29.28$ & $29.94$ / $28.76$ & $34.96$ / $34.86$ & $37.94$ / $35.82$ & $45.44$ / $44.53$ & $433$ / $779$ & $689$ / $1306$ \\  
            & fast  & $21.81$ / $20.92$ & $28.57$ / $28.20$ & $28.00$ / $26.66$ & $31.79$ / $31.94$ & $29.56$ / $28.02$ & $33.83$ / $33.76$ & $105$ / $186$ & $166$ / $306$ \\ \hline 

Mamba-2.8B-Zephyr & best  & $23.20$ / $22.09$ & $29.69$ / $29.37$ & $30.73$ / $28.84$ & $33.86$ / $34.11$ & $41.98$ / $37.95$ & $47.29$ / $45.85$ & $450$ / $793$ & $696$ / $1300$ \\  
            & fast  & $23.24$ / $21.72$ & $29.68$ / $29.01$ & $29.54$ / $27.32$ & $32.53$ / $32.47$ & $31.64$ / $28.91$ & $35.16$ / $34.45$ & $112$ / $191$ & $168$ / $308$ \\ \hline 

SmolLM-1.7B & best  & $22.44$ / $21.81$ & $29.59$ / $29.40$ & $30.31$ / $28.92$ & $34.80$ / $35.38$ & $39.26$ / $35.82$ & $46.10$ / $44.57$ & $249$ / $422$ & $374$ / $696$ \\  
            & fast  & $22.45$ / $21.03$ & $29.19$ / $28.41$ & $28.92$ / $27.00$ & $32.55$ / $32.66$ & $30.57$ / $28.51$ & $34.56$ / $34.36$ & $\mathbf{57}$ / $\mathbf{100}$ & $\mathbf{84}$ / $\mathbf{167}$ \\ \hline 

GPT2-XL (1.6B) & best  & $19.67$ / $12.06$ & $26.59$ / $16.91$ & $26.96$ / $15.84$ & $31.94$ / $20.26$ & $34.13$ / $19.80$ & $41.37$ / $25.79$ & $265$ / $453$ & $397$ / $833$ \\  
            & fast  & $19.58$ / $11.43$ & $25.96$ / $15.94$ & $25.31$ / $14.72$ & $29.28$ / $18.34$ & $26.84$ / $15.63$ & $31.31$ / $19.65$ & $\underline{62}$ / $\underline{107}$ & $\underline{96}$ / $\underline{180}$ \\ \hline 

Average & best  & $23.43$ / $22.03$ & $30.10$ / $29.26$ & $31.21$ / $28.85$ & $34.79$ / $34.31$ & $41.27$ / $36.77$ & $47.00$ / $44.49$ & $640$ / $1105$ & $983$ / $1922$ \\  
& fast  & $23.00$ / $21.01$ & $29.53$ / $28.12$ & $29.42$ / $26.61$ & $32.74$ / $31.80$ & $31.20$ / $28.08$ & $34.93$ / $33.66$ & $160$ / $275$ & $239$ / $452$ \\ 
\hline
             \hline
\end{tabular}}
  \caption{\label{table:results}
    Results comparing the performance and of the 9 evaluated models on both metrics for $k=1,3,k_{max}$, with \emph{best} and \emph{fast} configurations, each with mean and p90 latency. In each cell we report the results for both datasets: OpenAssistant / ShareGPT. For each metric and $k$, the winner is marked in bold and the second best is underlined.
  }
\end{table*}

\begin{figure*}[!h]
    \centering
    \includegraphics[width=0.85\textwidth]{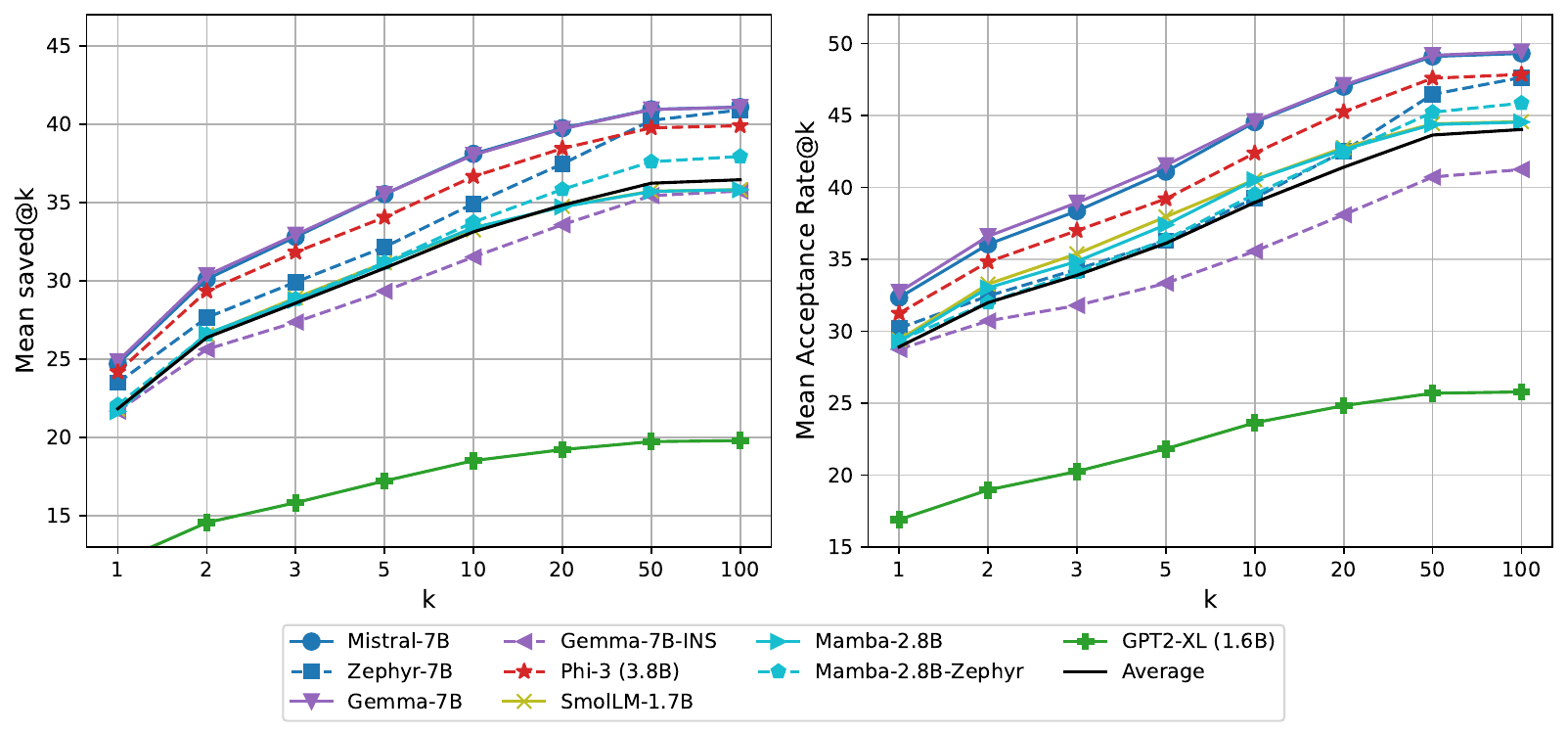}
    \caption{\emph{\textbf{saved@k}} and \emph{\textbf{acc. rate@k}} on ShareGPT for varying $k$ values.}
\label{fig:sgpt_saved_acc_rate@k}
\end{figure*}

\begin{figure}[!h]
\centering
    \includegraphics[width=0.85\columnwidth]{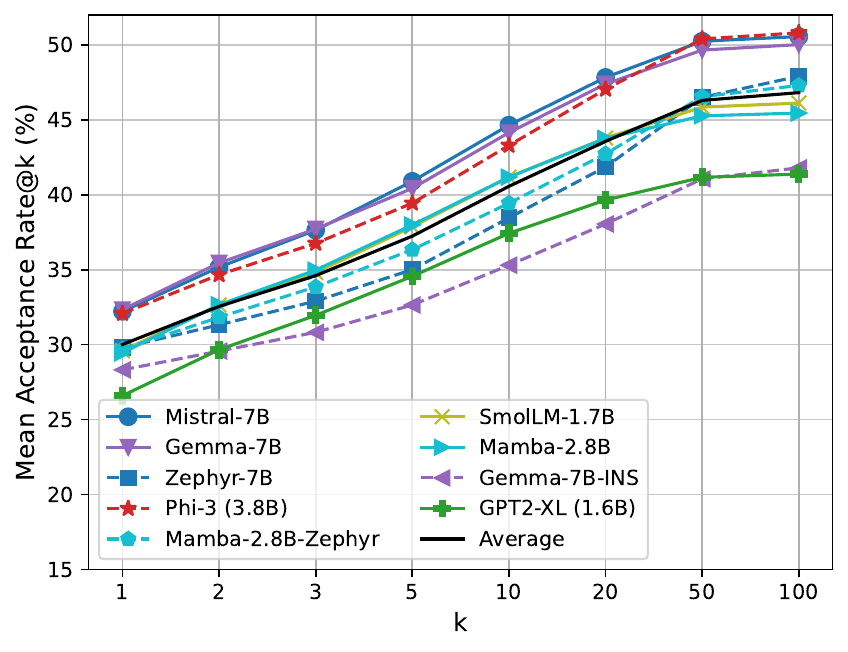}
    {\caption{\textit{\textbf{acc. rate@k}} on OASST for varying $k$ values.}\label{fig:acc_rate@k_vs_k}}
\end{figure}

Results on $k\text{@}1$ and $k\text{@}3$, representing realistic scenarios where the user is presented a single or $3$ completion suggestions, demonstrate that while current models are able to perform fairly on this task -- reaching acceptance rate of up to $\sim 37.5\%$, and saving the user the typing of up to $\sim 34.5\%$ of the characters --  there is still much room for improvement.
$k_{\text{max}}$ shows results considering all generated completions (at most, $n_c \times n_t$). While this scenario is not realistic, since for the \emph{best} configuration it means presenting the user with $100$ completion options, it shows the potential each solution has with a perfect ranking solution. The large gap between the $k_{\text{max}}$ results and the results with smaller $k$ values suggests that perplexity may be insufficient for ranking. This is in line with prior work \cite{selfcheckgpt, perplexity_ood, fact_checking_ppl}.


As for comparing the different models, the best performing model is Gemma-7B, which is also the model with the longest latency. Phi-3 stands out as well, with performance surpassing most of the other models, although they are larger in size and slower in latency. This result is consistent with its performance on other benchmarks compared to other models included in our evaluation \cite{phi-3}.
When comparing instruct models to their corresponding base models, instruct models mostly performed worse. This is likely due to the fact that the language modeling objective of the pretraining phase is closer to our task than the objective of the alignment phase. 

Finally, our \textit{best} vs \textit{fast} hyper parameter combinations are indeed able to offer a trade-off between latency and performance. On average, \textit{fast} is able to save $\sim 75\%$ of the latency compared to \textit{best}, while \textit{best} performs $\sim 30\%$ better on $k_{\text{max}}$ and $\sim 4-8\%$ better on the realistic $k$ scenarios.

Results on ShareGPT for varying $k$ values, complementing Figures \ref{fig:saved@k_vs_k} and \ref{fig:acc_rate@k_vs_k} in section \ref{subsec:benchmark} are shown in
Figure \ref{fig:sgpt_saved_acc_rate@k}.

  \begin{table*}[!t]
 \label{table:finetun}
  \centering
\resizebox{0.8\textwidth}{!}{
\begin{tabular}{lllllll} \hline \hline &      \multicolumn{3}{c}{\textbf{Mistral-7B}}                                     & \multicolumn{3}{c}{\textbf{Zephyr-7B}}                                  \\ 
&      \multicolumn{1}{c}{Pretrained} & \multicolumn{1}{c}{Finetuned} & \multicolumn{1}{c}{Improvement ($\%$)} & \multicolumn{1}{c}{Pretrained} & \multicolumn{1}{c}{Finetuned} & \multicolumn{1}{c}{Improvement ($\%$)}\\ \hline
 saved@$1$ & $25.99$ & $27.12$ & + $4.37$ & $24.03$ & $26.94$ & + $12.10$ \\
 acc. rate@$1$ & $32.22$ & $33.60$ & + $4.29$ & $29.80$ & $33.26$ & + $11.61$ \\\hline
 saved@$3$ & $34.68$ & $35.98$ & + $3.74$ & $31.08$ & $35.63$ & + $14.62$ \\
 acc. rate@$3$ & $37.64$ & $38.59$ & + $2.54$ & $32.91$ & $38.23$ & + $16.17$ \\\hline
 saved@$k_\text{max}$ & $44.89$ & $46.89$ & + $4.46$ & $44.03$ & $46.70$ & + $6.06$ \\
 acc. rate@$k_\text{max}$ & $50.56$ & $52.10$ & + $3.05$ & $47.91$ & $52.39$ & + $9.35$ \\\hline
             \hline
\end{tabular}}
  \caption{\label{table:finetun}
    saved@$k$ results of Mistral and Zephyr before and after fine-tuning.}
\end{table*}

\section{Fine-tuning Models to Improve AC}
\label{app:finetune}

We observe that fine-tuning models can offer further improvement upon the corresponding pre-trained models. We fine-tuned Mistral-7B and Zephyr-7B on the OASST train set using LoRA \cite{hu2021loralowrankadaptationlarge}, with the following hyperparameters (Mistral / Zephyr, respectively): learning rate $1.4\mathrm{e}{-4} / 2.4\mathrm{e}{-4}$, epochs $0.40 / 0.25$, batch size $16 / 16$.
In Table \ref{table:finetun} we report an average increase of $4.19\%$ and $10.93\%$ in the saved@$k$ metric for Mistral-7B and Zephyr-7B, respectively.

\section{Hyper Parameter Study} \label{sec:hp_study}

\begin{figure}[h]
\centering
    \includegraphics[width=7.5cm]{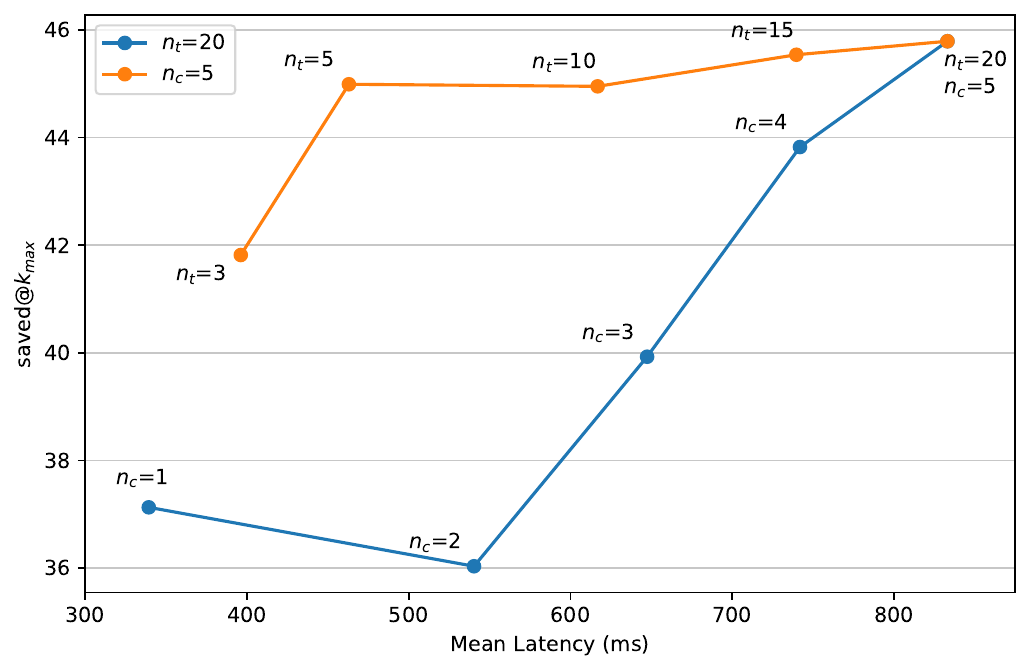}
    {\caption{\textbf{Hyper parameter study ($n_c$ and $n_t)$}. For each line, one of the parameters is fixed and the other is varied. Results are shown on the OASST dataset using the \textit{best} configuration and the Mistral model.}\label{fig:hp}}
\end{figure}

The auto completion method we use, extracting completion suggestions for language models, has two hyper parameters, $n_c$ and $n_t$, as detailed in Section \ref{models}. \\
In Figure \ref{fig:hp}, we show results on different values for the two parameters. In each figure, one of the parameters is fixed and the other is varied.

We also report results on different values of $k$ in Figures \ref{fig:acc_rate@k_vs_k} (acceptance rate) and \ref{fig:saved@k_vs_k} ($saved@k$). This parameter decides how many suggestions are shown to the user. While a higher value is guaranteed to increase the performance metrics, it may also incur slower latency and a cognitive cost for the user, and therefore for very high values it is unrealistic.

  \begin{table*}[!b]
 \label{table:main}
  \centering
\resizebox{1.0\textwidth}{!}{
\begin{tabular}{p{16cm}p{2cm}} \hline \hline 
Interaction & saved@$k_{max}$

(\%) \\ \hline

\color{lightgray}{<|prompter|> Do} \color{green}{\underline{you}}
\color{black}{have any information about} \color{green}\underline{the} \color{black}{Commodore} \color{green}\underline{64}?  
&
\color{black}{18.75}
\\
\hline

\color{lightgray}{<|prompter|> What is the Sun?

<|assistant|> The Sun is a star, a massive celestial body that converts matter into energy. It is the largest object in the solar system and is at its center; the gravitational field it exerts pulls the Earth, and all other planets and objects in the solar system, causing them to orbit it. The Sun produces energy by fusing hydrogen atoms in its core, producing helium. This continuous nuclear reaction emits tremendous amounts of energy, which is released as solar radiation. This radiation reaches the Earth and provides most of the energy needed to sustain life.

<|prompter|> Can} \color{green}{\underline{you}} \color{green}{\underline{tell me more about}}
\color{black}{suns} \color{green}\underline{from} \color{green}\underline{other} \color{green}\underline{solar systems?}  
&
\color{black}{83.02}
\\
\hline

\color{lightgray}{<|prompter|> Hi. Could you please tell me} \color{green}{\underline{how} \underline{I} \underline{can} \underline{get}} \color{black}{rid} \color{green}{\underline{of}} \color{black}{bad} \color{green}{\underline{breath?}}
&
\color{black}{59.38} 
\\
\hline

\color{lightgray}{<|prompter|> Can you tell me a bit about what has gone into your creation?

<|assistant|> My creation took over a month and the process of gathering data is still ongoing. I am becoming a larger assistant that is Open-Source.

<|prompter|> Do you have any way of keeping stats on how much data you've consumed or how long your training has taken?

<|assistant|> Yes but the public release of that information is pending review and validation.

<|prompter|> Here are 10 more question:} \color{green}{\underline{What}} \color{green}{\underline{kind of}} \color{black}{neural} \color{green}{\underline{network}} \color{black}{architecture was} \color{green}{\underline{used}} \color{green}{\underline{to}} \color{black}{create} \color{green}{\underline{you?}} 
&
\color{black}{42.19}
\\
\hline

\color{lightgray}{<|prompter|> I would like you to create some regex to find out if the first number in a set of numbers and letters is a one. For example, in “1 month, 2 days, and 3 hours” the selected} \color{green}{\underline{text would be}} \color{black}{"1."}
&
\color{black}{72.22}
\\
\hline

\color{lightgray}{<|prompter|> Write an "about us" paragraph for the website of a design company. The paragraph should contain information on the different services and strengths of the company. It is written in} \color{green}{\underline{a}} \color{black}{serious} \color{green}{\underline{tone and}} \color{black}{shouldn't feature} \color{green}{\underline{any technical}} \color{green}{\underline{terms}} \color{black}{\color{black}{that customers could be confused}} \color{green}{\underline{about.}}
&
\color{black}{34.38}
\\

 \hline 
             \hline
\end{tabular}}

  \caption{\label{table:full_interaction}{
    Full interaction examples with corresponding saved@k scores of randomly drawn prefixes from the OpenAssistant validation set. Completions were generated using Mistral-7B, with $n_c = 5, n_t=20$. The existing prefix including the context is colored gray. Accepted text is colored green with each acceptance underlined separately. Accepted text is colored green, and each accepted segment is separately underlined. If no suggestion was accepted, the text typed by the user is colored black.
    }
  }
\end{table*}

  \begin{table*}[ht]
 \label{table:examples}
  \centering
\resizebox{1.0\textwidth}{!}{
\begin{tabular}{p{2.4cm}|p{5.7cm}p{6.3cm}p{6.3cm}} \hline \hline
\textbf{Prefix}: & \multicolumn{3}{p{19cm}} {"<|prompter|> Who would win a chess match between Keanu Reeves and Audrey Hepburn?
<|assistant|> In a purely physical fight, Audrey would likely win. Keanu is renowned for being a very nice guy, I doubt he would hit a woman.  Audrey grew up in an era when women had to be serious badasses to get by.  She'd kick him in the nuts before he had time to say 'Vaya con Dios, Brah'
<|prompter|> No,"} \\
\textbf{Ground Truth}: & \multicolumn{3}{p{18cm}} {"I mean the board game chess!"} \\ \hline 
Mistral & I mean a chess match.  & I mean a chess  & I mean  \\ 
Zephyr & I meant a chess match.  & I meant a chess  & that's not what I meant.  \\ 
Phi & I  & I'm  & I'm asking for a chess match,  \\ 
Mamba & I mean  & I mean in a  & I  \\ 
GPT2 & Keanu  & I  & seriously, which of these men would win in a  \\ 
\hline \hline 

\textbf{Prefix}: & \multicolumn{3}{p{19cm}} {"<|prompter|> Hello, what education should I take at university? What are important"} \\
\textbf{Ground Truth}: & \multicolumn{3}{p{18cm}} {"things to consider when making the choice?"} \\ \hline 
Mistral & things to  & criteria in choosing a  & criteria in choosing a future profession and university?  \\ 
Zephyr & skills to  & skills  & educations in the future?  \\ 
Phi & skills to learn?  & classes to take in order to become a  & skills to develop?  \\ 
Mamba & courses?  & courses to  & courses to study  \\ 
GPT2 & book titles to read?  & subjects  & subjects and should I  \\ 
\hline \hline 

\textbf{Prefix}: & \multicolumn{3}{p{19cm}} {"<|prompter|> How do I code pong in python.
<|assistant|> Pong is a classic game that can be coded in Python using the pygame library. To get started, you'll need to install the pygame library. You can do this by running the command "pip install pygame" in your terminal. Once you have the library installed, you can create a new file and start coding. You'll need to create a window, draw the paddles, and create the ball. You'll also need to create the logic for the game, such as the ball bouncing off the paddles and the score keeping. Finally, you'll need to create the game loop that will keep the game running until the player wins or loses. With some practice and patience, you should be able to code a basic version of Pong in Python.
<|prompter|> Start writing code. Use pygame and make the window default to full screen and be resizable. There should be 2 modes,"} \\
\textbf{Ground Truth}: & \multicolumn{3}{p{18cm}} {"against computer and against human, these should be selected when a new game is created."} \\ \hline 
Mistral & single player and multiplayer.  & single player and  & single player  \\ 
Zephyr & single player  & single player and  & single  \\ 
Phi & one where the ball bounces off the  & one where the ball bounces off  & easy and hard.  \\ 
Mamba & one for the ball and one for the paddle.  & one for the ball and one for the  & one for the ball and one for  \\ 
GPT2 & fullscreen and windowed.  & windowed and full screen.  & fullscreen and  \\ 
\hline \hline 

\textbf{Prefix}: & \multicolumn{3}{p{19cm}} {"<|prompter|> write a inspirational monologue script from a spartan general telling his army that despite the impossible nature of their fight, that it is still worth fighting. do not directly reference sparta or the fight, but"} \\
\textbf{Ground Truth}: & \multicolumn{3}{p{18cm}} {"focus more on the concept of the indomitable human spirit and the will to keep fighting no matter the cost."} \\ \hline 
Mistral & it can be inferred  & it can be inferred (or outright stated) that the  & it can be inferred (or outright stated) that  \\ 
Zephyr & rather speak in general terms about perseverance  & rather  & rather speak in general terms about  \\ 
Phi & focus on themes of unity, courage, and  & focus on themes of  & focus on themes of unity, courage, and the  \\ 
Mamba & the gist is the same  & make a general  & instead the spirit of bravery and honor.  \\ 
GPT2 & do reference the  & instead  & do reference  \\ 
\hline \hline 

\textbf{Prefix}: & \multicolumn{3}{p{19cm}} {"<|prompter|> What are some unique, creative, and efficient ways to decorate and make the most of a small apartment space while still ensuring a comfortable living environment? Are there any particular design styles or techniques that are especially well-suited for small spaces, and what are the pros and cons of each approach? Are there any furniture pieces or items that are particularly useful for maximizing space and comfort in a small apartment, and what are"} \\
\textbf{Ground Truth}: & \multicolumn{3}{p{18cm}} {"some tips for choosing and arranging these items in a functional and aesthetically pleasing way?"} \\ \hline 
Mistral & some tips for choosing the right  & some  & some tips for choosing the right pieces for  \\ 
Zephyr & their benefits and drawbacks?  & some tips for arranging and organizing these items in a  & some tips for  \\ 
Phi & their benefits and drawbacks?  & some examples of these items?  & some examples of  \\ 
Mamba & the pros and cons of  & the pros and cons  & the pros and  \\ 
GPT2 & their pros and cons?  & their pros and  & their pros  \\ 
\hline \hline 
\end{tabular}}
  \caption{\label{table:examples} Comparison of top 3 suggested completions of different LLMs, on prefixes randomly drawn from the OpenAssistant validation set. Completions were generated with $n_c = 5, n_t=20$.
  }
\end{table*}

\begin{figure*}[h]
    \centering
    \includegraphics[width=1\textwidth]{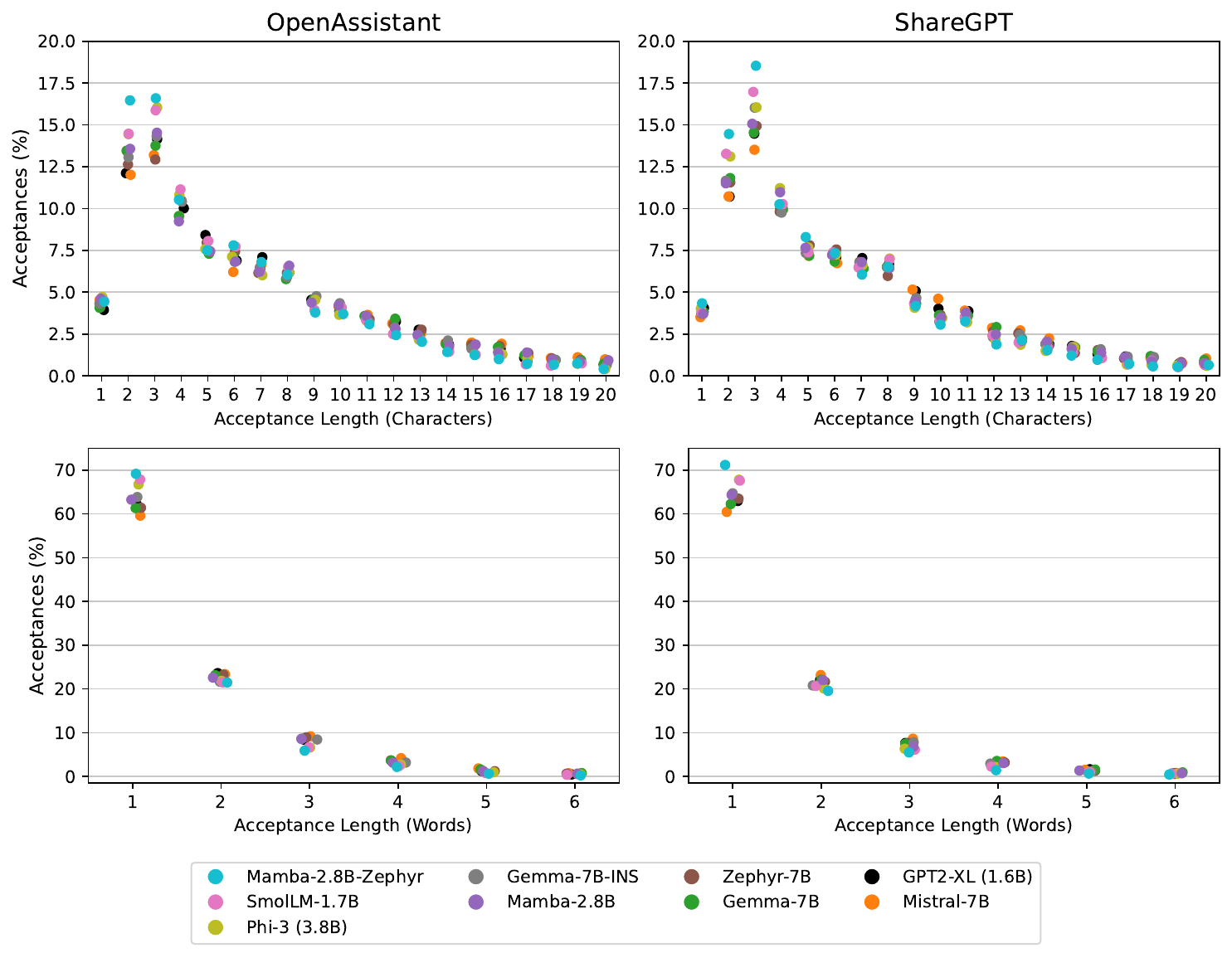}
    \caption{Lengths of accepted completions for $k=100$.}
\label{fig:length}
\end{figure*}

  \begin{table*}[ht]
 \label{table:main}
  \centering
\resizebox{1.0\textwidth}{!}{
\begin{tabular}{llccccccc} \hline \hline &      & \multicolumn{2}{c}{$k=1$}                                     & \multicolumn{2}{c}{$k=3$}                                     & \multicolumn{2}{c}{$k_{max}$}                                \\ 
&      & \multicolumn{1}{c}{saved@1} & \multicolumn{1}{c}{acc. rate@1} & \multicolumn{1}{c}{saved@3} & \multicolumn{1}{c}{acc. rate@3} & \multicolumn{1}{c}{saved@k} & \multicolumn{1}{c}{acc. rate@k} \\ \hline \hline

Mistral-7B & Single Word  & $26.00$ / $24.65$ & $42.42$ / $41.75$ & $34.70$ / $31.67$ & $54.01$ / $51.88$ & $36.02$ / $32.80$ & $55.86$ / $53.48$ \\  
          & EOS  & $15.35$ / $13.73$ & $12.60$ / $12.84$ & $19.47$ / $16.81$ & $16.00$ / $15.71$ & $19.87$ / $17.17$ & $16.33$ / $16.08$ \\ 
          & Partial  & $25.97$ / $24.67$ & $32.23$ / $32.32$ & $34.66$ / $32.76$ & $37.65$ / $38.35$ & $44.86$ / $41.04$ & $50.56$ / $49.33$ \\ 
 \hline 

Zephyr-7B & Single Word  & $25.63$ / $24.31$ & $41.76$ / $40.57$ & $33.89$ / $31.43$ & $52.87$ / $50.80$ & $35.00$ / $32.65$ & $54.43$ / $52.56$ \\  
          & EOS  & $11.95$ / $10.73$ & $9.53$ / $10.06$ & $15.85$ / $13.61$ & $12.46$ / $12.51$ & $16.91$ / $14.37$ & $13.35$ / $13.26$ \\ 
          & Partial  & $24.01$ / $23.47$ & $29.81$ / $30.22$ & $31.06$ / $29.89$ & $32.91$ / $34.31$ & $44.00$ / $40.85$ & $47.91$ / $47.65$ \\ 
 \hline 



Gemma-7B & Single Word  & $25.93$ / $24.53$ & $42.64$ / $41.70$ & $34.28$ / $31.75$ & $53.76$ / $51.87$ & $35.49$ / $32.77$ & $55.39$ / $53.29$ \\  
          & EOS  & $15.54$ / $13.78$ & $12.64$ / $12.95$ & $19.23$ / $17.04$ & $15.56$ / $15.91$ & $19.67$ / $17.51$ & $15.96$ / $16.37$ \\ 
          & Partial  & $25.80$ / $24.84$ & $32.34$ / $32.71$ & $34.66$ / $32.91$ & $37.72$ / $38.93$ & $44.75$ / $41.02$ & $50.02$ / $49.44$ \\ 
 \hline 

Gemma-7B-INS & Single Word  & $23.85$ / $22.51$ & $39.57$ / $38.32$ & $30.62$ / $28.21$ & $48.10$ / $46.20$ & $31.40$ / $28.62$ & $49.20$ / $46.84$ \\  
          & EOS  & $11.54$ / $10.20$ & $9.41$ / $9.68$ & $15.34$ / $13.19$ & $12.30$ / $12.29$ & $16.31$ / $13.70$ & $13.06$ / $12.76$ \\ 
          & Partial  & $22.08$ / $21.65$ & $28.33$ / $28.72$ & $28.13$ / $27.32$ & $30.83$ / $31.78$ & $38.94$ / $35.67$ & $41.81$ / $41.24$ \\ 
 \hline 

Phi-3 (3.8B) & Single Word  & $26.04$ / $24.12$ & $42.06$ / $40.55$ & $34.73$ / $31.17$ & $53.90$ / $50.62$ & $36.18$ / $32.20$ & $55.96$ / $52.22$ \\  
          & EOS  & $15.54$ / $12.58$ & $12.57$ / $11.89$ & $19.42$ / $15.38$ & $15.64$ / $14.47$ & $19.77$ / $15.67$ & $15.98$ / $14.78$ \\ 
          & Partial  & $26.07$ / $24.18$ & $32.07$ / $31.25$ & $34.42$ / $31.83$ & $36.76$ / $36.99$ & $45.18$ / $39.91$ & $50.81$ / $47.84$ \\ 
 \hline 

Mamba-2.8B & Single Word  & $22.19$ / $21.79$ & $37.52$ / $37.42$ & $29.72$ / $28.34$ & $48.07$ / $47.05$ & $31.04$ / $29.25$ & $49.85$ / $48.36$ \\  
          & EOS  & $12.92$ / $11.18$ & $10.90$ / $10.54$ & $15.79$ / $13.68$ & $13.34$ / $12.82$ & $16.05$ / $13.91$ & $13.62$ / $13.04$ \\ 
          & Partial  & $22.36$ / $21.66$ & $29.44$ / $29.28$ & $29.94$ / $28.76$ & $34.96$ / $34.86$ & $37.94$ / $35.82$ & $45.44$ / $44.53$ \\ 
 \hline 

Mamba-2.8B-Zephyr & Single Word  & $24.55$ / $22.81$ & $40.22$ / $38.71$ & $32.68$ / $29.46$ & $51.48$ / $48.54$ & $33.74$ / $30.51$ & $52.93$ / $50.04$ \\  
          & EOS  & $12.21$ / $10.47$ & $9.90$ / $9.59$ & $15.41$ / $12.72$ & $12.42$ / $11.64$ & $15.92$ / $13.05$ & $12.87$ / $12.03$ \\ 
          & Partial  & $23.20$ / $22.09$ & $29.69$ / $29.37$ & $30.73$ / $28.84$ & $33.86$ / $34.11$ & $41.98$ / $37.95$ & $47.29$ / $45.85$ \\ 
 \hline 

SmolLM-1.7B & Single Word  & $22.93$ / $21.90$ & $38.52$ / $37.56$ & $30.81$ / $28.22$ & $49.20$ / $46.97$ & $31.91$ / $29.00$ & $50.70$ / $48.11$ \\  
          & EOS  & $13.17$ / $11.82$ & $10.91$ / $11.31$ & $16.14$ / $14.08$ & $13.27$ / $13.45$ & $16.51$ / $14.45$ & $13.57$ / $13.83$ \\ 
          & Partial  & $22.44$ / $21.81$ & $29.59$ / $29.40$ & $30.31$ / $28.92$ & $34.80$ / $35.38$ & $39.26$ / $35.82$ & $46.10$ / $44.57$ \\ 
 \hline 

GPT2-XL (1.6B) & Single Word  & $20.07$ / $12.13$ & $34.16$ / $21.09$ & $26.72$ / $15.74$ & $43.65$ / $26.81$ & $27.94$ / $16.22$ & $45.33$ / $27.58$ \\  
          & EOS  & $11.38$ / $5.85$ & $9.52$ / $5.47$ & $13.56$ / $7.00$ & $11.37$ / $6.61$ & $13.77$ / $7.08$ & $11.58$ / $6.72$ \\ 
          & Partial  & $19.67$ / $12.06$ & $26.59$ / $16.91$ & $26.96$ / $15.84$ & $31.94$ / $20.26$ & $34.13$ / $19.80$ & $41.37$ / $25.79$ \\ 
 \hline 

Average & Single Word  & $24.10$ / $22.28$ & $40.06$ / $38.04$ & $31.97$ / $28.63$ & $50.76$ / $47.28$ & $33.12$ / $29.52$ & $52.34$ / $48.57$ \\  
          & EOS  & $12.30$ / $10.44$ & $10.04$ / $9.82$ & $15.91$ / $13.29$ & $12.94$ / $12.41$ & $16.47$ / $13.70$ & $13.42$ / $12.82$ \\ 
          & Partial  & $23.43$ / $22.03$ & $30.10$ / $29.26$ & $31.21$ / $28.85$ & $34.79$ / $34.31$ & $41.27$ / $36.77$ & $47.00$ / $44.49$ \\ 
 \hline
             \hline
\end{tabular}}

  \caption{\label{table:baselines_full}
    Scores of partial completions vs single word and full sentence baselines. OpenAssistant/ShareGPT.
  }
\end{table*}

\begin{table*}[ht]
  \centering
\resizebox{1\textwidth}{!}{
\begin{tabular}{ccccc|ccccc} \hline \hline
$n_c$ & $n_t$ & Hist. Len & saved@100 & Latency p90 (ms) & $n_c$ & $n_t$ & Hist. Len & saved@100 & Latency p90 (ms)\\ \hline
$5$ & $10$ & $Full$ & $45.75$ & $974$  & $3$ & $10$ & $1000$ & $38.56$ & $520$ \\
$5$ & $20$ & $Full$ & $45.60$ & $1287$  & $5$ & $5$ & $250$ & $38.32$ & $275$ \\
$5$ & $5$ & $Full$ & $45.00$ & $815$  & $4$ & $10$ & $250$ & $37.46$ & $419$ \\
$5$ & $20$ & $1000$ & $44.32$ & $947$  & $3$ & $3$ & $Full$ & $37.35$ & $468$ \\
$5$ & $5$ & $1000$ & $44.08$ & $451$  & $3$ & $3$ & $1000$ & $37.33$ & $278$ \\
$5$ & $10$ & $1000$ & $43.43$ & $614$  & $4$ & $20$ & $250$ & $36.58$ & $752$ \\
$4$ & $20$ & $Full$ & $43.13$ & $1137$  & $5$ & $3$ & $250$ & $36.42$ & $216$ \\
$4$ & $10$ & $Full$ & $43.02$ & $843$  & $4$ & $5$ & $250$ & $36.25$ & $254$ \\
$4$ & $10$ & $1000$ & $42.52$ & $569$  & $4$ & $3$ & $250$ & $34.14$ & $184$ \\
$4$ & $20$ & $1000$ & $42.40$ & $885$  & $3$ & $20$ & $250$ & $33.99$ & $732$ \\
$5$ & $3$ & $Full$ & $42.28$ & $742$  & $3$ & $10$ & $250$ & $33.37$ & $405$ \\
$4$ & $5$ & $Full$ & $41.81$ & $673$  & $3$ & $5$ & $250$ & $33.26$ & $241$ \\
$5$ & $20$ & $500$ & $41.45$ & $828$  & $5$ & $20$ & $100$ & $32.59$ & $732$ \\
$5$ & $3$ & $1000$ & $41.10$ & $388$  & $3$ & $3$ & $250$ & $32.13$ & $171$ \\
$4$ & $5$ & $1000$ & $40.85$ & $399$  & $5$ & $5$ & $50$ & $25.35$ & $219$ \\
$3$ & $10$ & $Full$ & $40.44$ & $695$  & $5$ & $20$ & $50$ & $25.21$ & $729$ \\
$3$ & $20$ & $Full$ & $40.35$ & $991$  & $5$ & $10$ & $50$ & $24.95$ & $393$ \\
$4$ & $3$ & $Full$ & $39.67$ & $602$  & $5$ & $3$ & $50$ & $23.45$ & $148$ \\
$3$ & $20$ & $1000$ & $39.59$ & $818$  & $4$ & $10$ & $50$ & $23.01$ & $389$ \\
$3$ & $5$ & $Full$ & $39.57$ & $526$  & $4$ & $20$ & $50$ & $22.94$ & $723$ \\
$5$ & $20$ & $250$ & $39.47$ & $776$  & $4$ & $5$ & $50$ & $22.67$ & $216$ \\
$5$ & $10$ & $250$ & $38.80$ & $435$  & $4$ & $3$ & $50$ & $22.07$ & $147$ \\
$4$ & $3$ & $1000$ & $38.76$ & $330$  & $3$ & $10$ & $50$ & $21.02$ & $385$ \\
$3$ & $5$ & $1000$ & $38.59$ & $348$  & $3$ & $20$ & $50$ & $20.90$ & $717$ \\
\hline \hline
\end{tabular}}
  \caption{
    \textbf{Latency-Performance Trade-Off.} Full results for all configurations, complementing Table \ref{table:latency} in section \ref{sec:further_analysis}. 
    Mistral-7b evaluated on 
    the OASST test set. $n_c \in \{3,4,5\}$, $n_t \in \{3,5,10,20\}$, and context length $len(C) \in \{50, 250, 1000, Full\}$ (measured in characters). In total, 48 hyper-parameter configurations were evaluated. Results are sorted by their saved@100 score.
  }\label{table:latency_full}
\end{table*}
\end{document}